\documentclass[sigconf,review=false]{acmart}
\usepackage{multirow, adjustbox, booktabs, colortbl}
\usepackage{amsfonts}
\usepackage{amsmath}
\usepackage{natbib}
\usepackage{graphicx}
\usepackage{enumitem}
\usepackage{caption}
\usepackage{subcaption}
\usepackage{array}
\AtBeginDocument{%
  }

\copyrightyear{2025}
\acmYear{2025}
\setcopyright{rightsretained}
\acmConference[WWW Companion '25]{Companion Proceedings of the ACM Web Conference 2025}{April 28-May 2, 2025}{Sydney, NSW, Australia}
\acmBooktitle{Companion Proceedings of the ACM Web Conference 2025 (WWW Companion '25), April 28-May 2, 2025, Sydney, NSW, Australia} \acmDOI{10.1145/3701716.3715564} \acmISBN{979-8-4007-1331-6/2025/04}

\begin{document}

\title{Optimal Return-to-Go Guided Decision Transformer for Auto-Bidding in Advertisement}

\author{Hao Jiang}
\authornote{Both authors contributed equally to this research.}
\email{jianghao10@kuaishou.com}
\affiliation{%
  \institution{Kuaishou Technology}
  \city{Beijing}
  \country{China}
}

\author{Yongxiang Tang}
\authornotemark[1]
\email{tangyongxiang@kuaishou.com}
\affiliation{%
  \institution{Kuaishou Technology}
  \city{Beijing}
  \country{China}
}

\author{Yanxiang Zeng}
\email{zengyanxiang@kuaishou.com}
\affiliation{%
  \institution{Kuaishou Technology}
  \city{Beijing}
  \country{China}
}

\author{Pengjia Yuan}
\email{yuanpengjia@kuaishou.com}
\affiliation{%
  \institution{Kuaishou Technology}
  \city{Beijing}
  \country{China}
}

\author{Yanhua Cheng}
\email{chengyanhua@kuaishou.com}
\affiliation{%
  \institution{Kuaishou Technology}
  \city{Beijing}
  \country{China}
}

\author{Teng Sha}
\email{shateng@kuaishou.com}
\affiliation{%
  \institution{Kuaishou Technology}
  \city{Beijing}
  \country{China}
}

\author{Xialong Liu}
\email{zhaolei16@kuaishou.com}
\affiliation{%
  \institution{Kuaishou Technology}
  \city{Beijing}
  \country{China}
}

\author{Peng Jiang}
\orcid{0000-0002-9266-0780}
\email{jp2006@139.com}
\affiliation{%
  \institution{Kuaishou Technology}
  \city{Beijing}
  \country{China}
}

\ccsdesc[500]{Computing methodologies~Machine learning algorithms}
\ccsdesc[500]{Computing methodologies~Planning for deterministic actions}
\renewcommand{\shortauthors}{Hao Jiang et al.}
\begin{abstract}
In the realm of online advertising, advertisers partake in ad auctions to obtain advertising slots, frequently taking advantage of auto-bidding tools provided by demand-side platforms. To improve the automation of these bidding systems, we adopt generative models, namely the Decision Transformer (DT), to tackle the difficulties inherent in automated bidding. Applying the Decision Transformer to the auto-bidding task enables a unified approach to sequential modeling, which efficiently overcomes short-sightedness by capturing long-term dependencies between past bidding actions and user behavior. Nevertheless, conventional DT has certain drawbacks: (1) DT necessitates a preset return-to-go (RTG) value before generating actions, which is not inherently produced; (2) The policy learned by DT is restricted by its training data, which is consists of mixed-quality trajectories. To address these challenges, we introduce the $R^*$ Decision Transformer ($R^*$ DT), developed in a three-step process: (1) $R$ DT: Similar to traditional DT, $R$ DT stores actions based on state and RTG value, as well as memorizing the RTG for a given state using the training set; (2) $\hat R$ DT: We forecast the highest value (within the training set) of RTG for a given state, deriving a suboptimal policy based on the current state and the forecasted supreme RTG value; (3) $R^*$ DT: Based on $\hat R$ DT, we generate trajectories and select those with high rewards (using a simulator) to augment our training dataset. This data enhancement has been shown to improve the RTG of trajectories in the training data and gradually leads the suboptimal policy towards optimality. Comprehensive tests on a publicly available bidding dataset validate the $R^*$ DT's efficacy and highlight its superiority when dealing with mixed-quality trajectories.
\end{abstract}

\keywords{Offline Reinforcement Learning, Decision Transformer, Auto Bidding}

\maketitle

\begin{figure*}[t]
    \centering
    \includegraphics[width=0.75\textwidth]{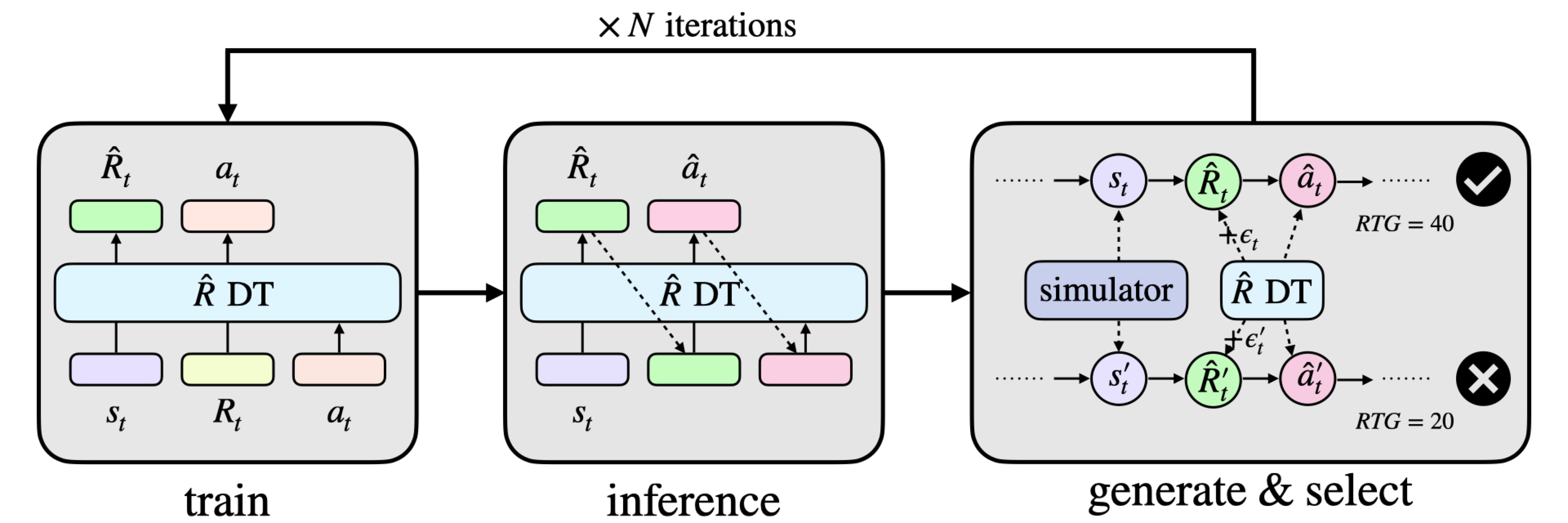}
    \caption{Our $R^*$ Decision Transformer Framework}
    \label{fig:framework}
    \vspace{-2mm}
\end{figure*}

\section{Introduction}
Auto-bidding poses a sophisticated optimization problem\cite{aggarwal2024auto,aggarwal2019autobidding,balseiro2021robust,balseiro2023joint,susan2023multi}, where the goal is to maximize conversions while staying within budgetary and cost limitations. Recent developments have highlighted the Decision Transformer (DT)\cite{chen2021decision} as an effective method for tackling sequential decision-making tasks, including auto-bidding. DTs utilize transformer capabilities\cite{parisotto2021efficient, esslinger2022deep,upadhyay2019transformer} to incorporate past states, actions, and rewards into a contextualized format, which enhances learning from offline datasets in reinforcement learning scenarios. This quality makes DTs especially apt for the sequential decision-making aspects of auto-bidding. The standard expression for DT is:
$$
    a_t(s_t, R_t) = {\rm DT}(s_{\leq t}, R_{\leq t}, a_{<t}),
$$
where $a_t$, $s_t$, and $R_t$ denote the action, state, and return-to-go at time $t$, respectively. The future rewards, $R_t$, is the cumulative sum of rewards following action $a_t$: $R_t=\sum_{i>t}r_i$. During training, $R_t$ is determined by the actual observed return values, whereas during testing, it is predefined.

As we demonstrated, DT in the auto-bidding problem comes with its own set of challenges. Two significant concerns include the issues of presetting the return-to-go\cite{lee2022multi} and the problem of selecting suboptimal trajectories\cite{wu2024elastic, janner2021offline}. First, the concept of Return-to-Go (RTG) is fundamental for DTs, as it guides the DT to take good actions or bad actions. Therefore, providing explicit RTG is essential for DTs to master policies aligned with specific objectives or rewards. However, manually assigned RTG values can lead to significant errors due to human inaccuracies or insufficient knowledge about the dynamics and reward system of the environment. Such mistakes might cause the model to learn suboptimal policies, training on incorrect targets and thereby straying from the ideal policy. Second, the Insufficient Trajectory Problem significantly impacts the optimization of bidding strategies in dynamic settings\cite{levine2020offline}. DTs that only utilize observed trajectories have built-in limitations when aiming to create globally optimal bidding paths. This issue occurs because observed data may not cover all possible system scenarios or states, leading to a bias toward common situations\cite{wang2022bootstrapped}. Consequently, the model's generalization capacity and predictive accuracy in novel or infrequent circumstances are reduced. The lack of exposure to unexplored trajectories may result in suboptimal decision-making, with DTs potentially failing to explore the full range of actions that could yield better outcomes. To address these challenges, it is necessary to devise techniques that allow DTs to more effectively use and learn from unobserved or simulated trajectories, thereby increasing their resilience and flexibility in complex, evolving settings.

To tackle the challenges previously discussed, we undertake three steps. Initially, we propose a new sequence structuring method known as the $R$ DT, which reorganizes the input sequence to include the state ($s$), the Return-to-Go ($R$), and the action ($a$). In this arrangement, the network predicts the RTG value, negating the requirement for a defined RTG. Subsequently, we introduce the $\hat R$ DT, which is designed to determine a suboptimal $R$ value that helps steer the policy towards optimal paths. This is achieved by predicting the RTG's upper bound for a given state through a loss function linked to quantile regression. Additionally, the use of the $\hat R$ DT enables the creation of new, high-quality trajectories that offer significant rewards, thereby increasing the proportion of high-quality data in the training set, aiding in the prediction of RTG's upper bound and the resulting policy. This data augmentation process is reiterated over several iterations, culminating in the refined DT, dictated by nearly optimal RTG values, termed as $R^*$ DT. Finally, experiments have been conducted to validate the proposed approach.

In summary, our primary contributions are as follows:
\begin{itemize} 
\item Our approach improves the Decision Transformer's input sequence by eliminating the need for user-specified Return-to-Go (RTG) values. Instead, the model inherently deduces RTG directly from the state. 
\item We introduce a new $\hat R$ loss function aimed at explicitly modeling the optimal RTG within the training data, guiding the model to learn high RTG values and the associated effective policies. 
\item We employ the $R^*$ Decision Transformer to produce and select high-quality trajectories, thereby refining decision-making through a broader exploration of examples to ultimately discover the optimal policy. 
\end{itemize}

\section{Methodology}





\subsection{Memorizing Actions and RTGs ($\boldsymbol{R}$ DT)}
We implement the decision transformer framework to memorize the action at time $t$ through $a_t = DT(s_{\leq t}, R_{\leq t})$, where $s_{\leq t}$ and $R_{\leq t}$ denote the states and cumulative future rewards from timestamp $0$ to $t$. Unlike the traditional DT model, we incorporate the memorization of RTG values for each state within the same transformer structure, referring to this improved framework as $R$ DT, and expressed as:
$$
    \begin{aligned}
        R_t (s_t) &= R{\rm \_DT}(s_{\leq t}, R_{<t}, a_{<t})
        \\ a_t (s_t, R_t)&= R{\rm \_DT}(s_{\leq t}, R_{\leq t}, a_{<t}).
    \end{aligned}
$$
The predicted RTG $R_t (s_t)$ is learned using the MSE loss. This adaptation allows the DT model to directly forecast an RTG value. However, this RTG value might correspond to a low-quality trajectory, potentially deteriorating the subsequent action generation. A straightforward approach is to avoid replicating RTG values in training samples and instead predict the maximum RTG value that fits a particular state within the training set. This strategy gives rise to the $\hat R$ DT discussed in the next section.

\subsection{Sorting Optimal RTG ($\boldsymbol{\hat R}$ DT)}
Given that the optimal policy $a_t$ at time $t$ is directed towards maximizing the future sum of reward, i.e. $R_t$, the following two-step approach can be adopted: first, estimate the optimal $R_t$ given the state $s_t$; second, forecast the action $a_t$ to fulfill this optimal $R_t$. Constrained in the scope of the training trajectories, the optimal $R_t$ conditioned on state $s_t$ is denoted as $\hat R_t$, while in all possible trajectories, the optimal RTG is represented as $R^*_t$. Hence, the inequality holds as follows:
\begin{equation}
    R^*_t(s_t)\geq {\hat R}_t(s_t)=\underset{s_t,R_T\sim \mathcal T}{\sup} R_t (s_t) \geq R_t (s_t),
    \label{q_in_eq}
\end{equation}
where $\sup$ is taken for $s_t$ and $R_t(s_t)$ from the training trajectories set $\mathcal T$. The estimation of $\hat R_t$ is expressed by:
$$
       \hat R_t (s_t) = \hat R {\rm \_DT}(s_{\leq t}, R_{<t}, a_{<t}),
$$
and the loss function utilized in the learning of $\hat R_t$ is:
$$
\mathcal L \left(\hat R_t(s_t)\right)=(1-\lambda){\rm ReLU}\left(R_t- \hat R_t(s_t) \right)+\lambda {\rm ReLU}\left(\hat R_t(s_t)-R_t  \right).
$$
This loss function is derived from quantile regression\cite{koenker2005quantile}, resembling the expectile regression used in IQL\cite{kostrikov2021offline}. For $\lambda \in (0,1)$, the solution corresponds to the $(1-\lambda)$th quantile of $R_t(s_t)$. As $\lambda$ approaches $0$, ${\hat R}_t(s_t)$ effectively approximates the maximum $R_t$ conditioned on $s_t$. Thus, $\hat{R}_t$ functions as a robust estimator of the upper limit of the return-to-go value.

\subsection{Applying and Revising Optimal Policy ($\boldsymbol{{R}^*}$DT)}
As we derive the suboptimal RTG estimator $\hat R_t$, we can derive the corresponding suboptimal policy informed by $\hat R_t$:
$$
    \hat a_t(s_t, \hat R_t) = \hat R {\rm \_DT}(s_{\leq t}, [R_{<t},\hat R_t], a_{<t}).
$$
However, this policy guided by suboptimal RTG presents challenges: (i) both $\hat R_t$ and $\hat a_t$ are affected by inaccuracy due to the rarity of high-quality state-reward pairs $(s_t, R_t)$ in the training set; (ii) the available training data may include low-quality trajectories that contribute to the subpar $\hat R_t$, which significantly deviates from the optimal RTG $R^*_t$. To address these issues, we enhance the training set by creating trajectories from the derived suboptimal policy $\hat a_t$:
\begin{equation}
    \begin{aligned}
        \hat R_t(s_t) &= \hat R {\rm \_DT}(s_{\leq t}, R_{<t}, \hat a_{<t}) + \epsilon_t
        \\ \hat a_t(s_t, \hat R_t) &= \hat R {\rm \_DT}(s_{\leq t}, [R_{<t},\hat R_t], \hat a_{<t}) 
        \\  R_t, s_{t+1} &= {\rm simulator}(s_t, \hat a_t).
    \end{aligned}
    \label{gen_tra}
\end{equation}
In this paper, we present a simulator for assessing generated trajectories. Within the auto-bidding framework, the simulator models the auction environment using opponents' eCPM data, which is obtainable from log records. Compared with competitors' eCPMs, we can deduce the advertisement's auction success based on the bid provided by the $\hat R$ DT. Furthermore, sampling from ${\mathcal Bernoulli}(pExposure)$ and ${\mathcal Bernoulli}(pCVR)$ allows us to simulate whether the advertisement is displayed and if it results in a conversion. This enables the state of the advertiser to be updated post-auction and facilitates the computation of RTG once the complete trajectory is produced.

The trajectory generation process adheres to the suboptimal policy and introduces diversity through stochastic noise from $\epsilon_t$. This process selects new trajectories based on their total RTG values, keeping those with higher RTG. We repeat the cycle of training, generating, and selecting, and denote the resulting training set at each stage as $\mathcal T^{(k)}$. In each iteration, the estimated suboptimal RTG and action are denoted as $\hat R^{(k)}_t$ and $\hat a^{(k)}_t$, respectively. Given that the sequence of training sets satisfies $\mathcal T^{(1)}\subset \cdots \subset \mathcal T^{(k)}$, and by the definition of $\hat R^{(k)}_t$, we observe $\hat R^{(1)}_t<\cdots <\hat R^{(k)}_t<R^*_t$. This methodology paves the way towards approximating $R^*_t$. Furthermore, with the continuous inclusion of high-quality trajectories in the training set, the issue of scarcity in high-quality state-reward pairs $(s_t, R_t)$ is significantly alleviated, allowing for more robust training of the suboptimal policy $\hat a_t$. After successive rounds of training set augmentation, the resulting decision transformer is signified as the $R^*$ DT. This approach thoroughly explores the upper bound of RTG, guiding the policy with near-optimal RTG values and approximates the optimal policy.

\section{Experiments}
\subsection{Experimental Setup}
\subsubsection{Datasets.}
We evaluate our $R^*$ decision transformer using a simulated bidding dataset \cite{AIGB_data}. This dataset comprises aggregated information from 48 advertisers participating in bids across 48 time points, capturing the (state, action, reward, subsequent state) data of various advertisers for each decision step within different delivery cycles. Furthermore, we enhance the dataset by altering each advertiser's target CPA, thereby creating three distinct datasets: the initial dataset (dataset 1), a high-target-CPA dataset (dataset 2), and a low-target-CPA dataset (dataset 3).

\subsubsection{Metrics.}
The three metrics we utilize are the average score, average conversion, and average budget consumption ratio. The primary goal is the average score, which is defined in terms of conversion and budget consumption ratio as detailed below:
$$\begin{aligned}
    &b_i=R^*\_{\rm DT}(s_i),
&\text { score }=\mathbb{P}(C_{a} ; C_{t})  \sum_i w_i  e_i  v_i,
\end{aligned}$$
where $b_i$ represents the bid action selected by $R^*$ DT. The variable $w_i=w(b_i,s_i)$ determines if the advertiser succeeds in winning the auction at a bid of $b_i$ conditioned on the state $s_i$. The term $e_i=e(b_i,s_i)$ indicates whether the ad is exposed, while $v_i=v(b_i,s_i)$ indicates whether a conversion occurs after the advertisement is exposed. The penalty function for exceeding the CPA limit is given by $\mathbb P(C_{a};C_{t})$ as follows:
$$\begin{aligned}
&C_{a}=\frac{\sum_i b_i w_i  e_i }{\sum_i w_i  e_i v_i},
&\mathbb{P}(C_{a} ; C_{t})=\min \left\{\left(\frac{C_{a}}{ C_{t}}\right)^2, 1\right\},
\end{aligned}$$
where the advertiser's target CPA is $ C_{t}$  and the actual CPA is $ C_{a}$.

\subsubsection{Baselines.}
We select three baseline approaches: the decision transformer (DT) as a sequential method; alongside behavior cloning (BC) and implicit Q-learning (IQL) as non-sequential offline RL techniques.

\subsection{Performance Comparison}
\begin{table}[h]
\small
\renewcommand\arraystretch{1.15}
\centering
\setlength{\tabcolsep}{0.6mm}{
\begin{tabular}{c|ccc|ccc|ccc}
\toprule
\multirow{2}*{Method} & \multicolumn{3}{c|}{Test Dataset 1} & \multicolumn{3}{c|}{Test Dataset 2} & \multicolumn{3}{c}{Test Dataset 3} \\
\cline{2-10}
& Score & Conv & Budget & Score & Conv & Budget & Score & Conv & Budget \\
\midrule
DT& 28.04 & 34.33 & 97.43\% & 29.45 & 34.43 & 98.15\% & 26.34 & 29.78 & 95.43\% \\
BC & 31.80 & 34.29 & 75.20\% & 33.02 & 34.58 & 75.12\% & 29.14 & 29.70 & 73.40\% \\
IQL & 34.10 & 36.29 & 81.95\% & 33.69 & 35.10 & 69.66\% & 29.27 & 30.20 & 72.32\% \\
$R^*$ DT & \textbf{36.97} & 41.89 & 97.97\% & \textbf{38.69} & 42.89 & 96.77\% & \textbf{35.00} & 37.81 & 78.19\% \\
\bottomrule
\end{tabular}}
\caption{Performance Comparison on Three Datasets}
\label{tab-1}
\vspace{-2em}
\end{table}
Table \ref{tab-1} presents the results across three datasets, where the result of the original DT is derived by experimenting with various initial RTG inputs. The table emphasizes the improved performance of the $R^*$ DT in terms of average score and conversion metrics, demonstrating its capability in overcoming the challenges associated with auto-bidding under target CPA constraints. Conversely, the base DT does not outperform the non-sequential BC and IQL methods, attributed to its shortcomings in determining the initial RTG value and choosing suboptimal trajectories. These issues are addressed by our $R^*$ DT approach. In various target CPA scenarios, $R^*$ DT effectively manages advertisers' budgets. Specifically, in the high-target-CPA case of dataset 2, it prioritizes optimizing the conversion component in the score function, whereas in the low-target-CPA case of dataset 3, it focuses on minimizing the target CPA penalty. As a result, it demonstrates robustness under different auction constraints within the auto-bidding task.

\subsection{Ablation Study}
\begin{table}[h]
\small
\renewcommand\arraystretch{1.15}
\centering
\setlength{\tabcolsep}{0.6mm}{
\begin{tabular}{c|ccc|ccc|ccc}
\toprule
\multirow{2}*{Method} & \multicolumn{3}{c|}{Test Dataset 1} & \multicolumn{3}{c|}{Test Dataset 2} & \multicolumn{3}{c}{Test Dataset 3} \\
\cline{2-10}
& Score & Conv & Budget & Score & Conv & Budget & Score & Conv & Budget \\
\midrule
$R$ DT & 32.68 & 36.00 & 84.37\% & 33.39 & 35.77 & 73.25\% & 30.08 & 31.64 & 66.60\% \\
$\hat R$ DT & 35.42 & 40.60 & 98.33\% & 33.41 & 35.79 & 73.34\% & 31.55 & 32.89 & 66.27\% \\
$R^*$ DT(1) & 34.34 & 37.20 & 84.40\% & 35.77 & 40.75 & 96.10\% & 32.25 & 34.16 & 70.09\% \\
$R^*$ DT(2) & 35.65 & 37.93 & 84.30\% & 36.55 & 41.12 & 93.82\% & 33.42 & 36.25 & 75.56\% \\
$R^*$ DT(3) & 36.61 & 38.68 & 84.20\% & 37.34 & 40.80 & 89.50\% & 33.67 & 36.33 & 73.97\% \\
$R^*$ DT(4) & 36.97 & 41.89 & 97.97\% & 37.98 & 42.27 & 97.46\% & 34.93 & 37.75 & 78.05\% \\
$R^*$ DT(5) & \textbf{36.97} & 41.89 & 97.97\% & \textbf{38.69} & 42.89 & 96.77\% & \textbf{35.00} & 37.81 & 78.19\% \\
\bottomrule
\end{tabular}}
\caption{Ablation Study}
\label{tab-2}
\vspace{-2em}
\end{table}
Table \ref{tab-2} showcases an in-depth ablation study assessing the performance of the proposed $R$ DT, $\hat R$ DT, and $R^*$ DT. We also examined the variation in performance across different iteration rounds for $R^*$ DT. As indicated by the test results, the performance of $R$ DT, $\hat R$ DT, and $R^*$ DT improves progressively, aligning with our theory. Examining various iteration rounds reveals that augmenting the training dataset through sample generation and selection by the simulator proves beneficial.

\subsection{Discussion on Predicted RTG}
\begin{figure}[h]
    \centering
    \includegraphics[width=0.36\textwidth]{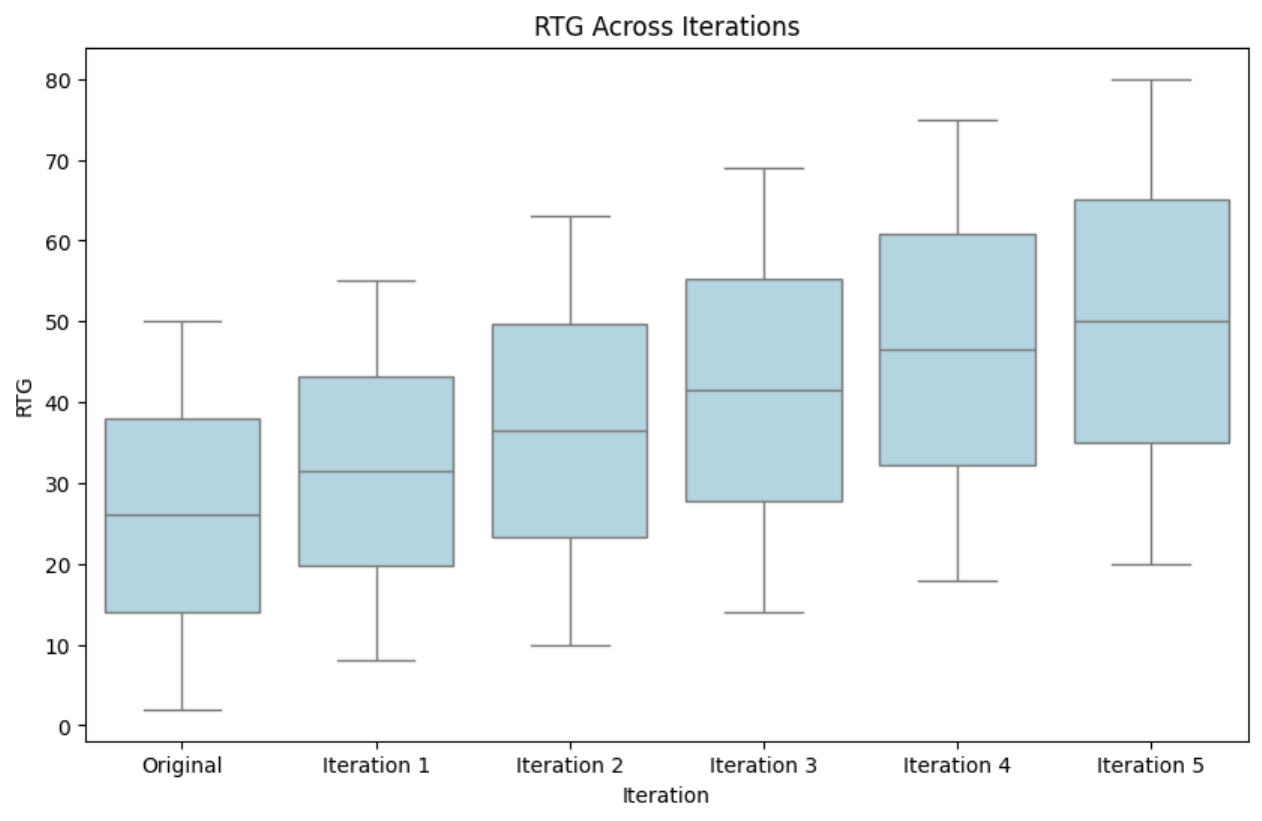}
    \caption{RTG (via simulator) performance of trajectories generated by the $R^*$ decision transformer.}
    \label{fig:rtg}
    \vspace{-2mm}
\end{figure}
Figure \ref{fig:rtg} illustrates a sequence of box-plots showcasing the consistent advancement in return-to-go (RTG) metrics across various iterations. Initially, the RTG values span from 2 to 50, but each subsequent iteration leads to a distinct upward trend of the RTG distribution. By the time the fifth iteration is reached, the RTG range extends from 20 to 80. This continual increase in RTG values with each iteration highlights the effectiveness of the iterative process in refining the decision-making strategy, allowing the $R^*$ DT to explore and capitalize on more optimal trajectories. Moreover, the test results in Table \ref{tab-2} show that the test results correlate with the improvement of RTG in the generated dataset.

\section{Conclusion}
This research presents the $R^*$ DT methodology to tackle the auto-bidding challenge in online advertising. The $R^*$ DT is progressively developed, beginning with improving the decision transformer (DT) via $R$ DT with RTG memorization. It then advances to the $\hat R$ DT, which employs suboptimal RTG prediction trained with the $\hat R$ loss. Finally, the $R^*$ DT is trained on an expanded dataset generated by the $\hat R$ DT and further refined through a simulator. This approach addresses issues related to predefining the return-to-go value and the sparsity of high-quality trajectories in the training dataset. The experimental findings reveal that our framework outperforms the DT, BC, and IQL methods in auto-bidding tasks, and the ablation study validates the effectiveness of the $\hat R$ loss and the iterative process of train sample generation and selection.

\bibliographystyle{ACM-Reference-Format}
\bibliography{sample-base}

\end{document}